\pdfoutput=1

\documentclass[11pt]{article}

\usepackage{naacl2021}

\usepackage{times}
\usepackage{latexsym}
\usepackage{amsmath}
\usepackage{tabularx}
\usepackage{xcolor}
\usepackage{booktabs}
\usepackage{amssymb}

\usepackage[T1]{fontenc}

\usepackage[utf8]{inputenc}

\usepackage{microtype}
\usepackage{ulem}

\renewcommand{\emph}[1]{\textit{#1}}

\title{\vspace*{-0.5in}
{
{\small \hfill NAACL'21}\\
\vspace*{.25in}}Improving Faithfulness in Abstractive Summarization \\ with Contrast Candidate Generation and Selection }

\author{Sihao Chen$^1$\thanks{\hspace{0.2cm}Most of the work done while the authors were at Google.} \hspace{1cm} Fan Zhang$^2$ \hspace{1cm} Kazoo Sone$^2$ \hspace{1cm} Dan Roth$^{1*}$ \\
$^1$University of Pennsylvania \hspace{1cm} $^2$Google \\
\small \texttt{\{sihaoc, danroth\}@cis.upenn.edu, \{zhanfan, sone\}@google.com} \\}

\date{}

\begin{document}
\maketitle
\begin{abstract}
Despite significant progress in neural abstractive summarization, 
recent studies have shown that the current models are prone to generating summaries that are \emph{unfaithful} to the original context. To address the issue, we study \emph{contrast candidate generation} and \emph{selection} as a model-agnostic post-processing technique to correct the extrinsic hallucinations (i.e. information not present in the source text) in unfaithful summaries. We learn a discriminative correction model by 
generating alternative candidate summaries where  named entities and quantities in the generated summary are replaced with ones with compatible semantic types from the source document. This model is then used to
select the best candidate as the final output summary. Our experiments and analysis across a number of neural summarization systems show that our proposed method is effective in identifying and correcting extrinsic hallucinations.
We analyze the typical hallucination phenomenon by different types of neural summarization systems, in hope to provide insights for future work on the direction.

\end{abstract}

\section{Introduction}


Abstractive Summarization is the task of producing a concise and fluent summary that is salient and \emph{faithful} to the source document(s). Data-driven, neural methods \cite{rush2015neural, nallapati2016abstractive, see2017get}, and the more recent, pretrained transformer language models \cite{vaswani2017attention, devlin2019bert, liu2019text}, have shown improvements in the fluency and salience of generated summaries.

However, less progress has been made on improving the \emph{faithfulness} of the generated summaries, that is, producing a summary that 
is \emph{entailed} by the information presented in the source document. 
Despite the increased level of performance under automatic metrics such as \textsc{Rouge} \cite{lin-2004-rouge} or \textsc{BertScore} \cite{bert-score},
current state of the art models~\cite{liu2019text, lewis2019bart} 
produce summaries that
suffer from \emph{intrinsic} and \emph{extrinsic hallucinations} -- 
the fabrication of untruthful text spans containing information either \emph{present} or \emph{absent} from the source \cite{Maynez2020OnFA}. 

\begin{table}
\small
\centering
\begin{tabular}{p{0.95\linewidth}}
\hline \textbf{Source}: He was re-elected for a second term by the UN General Assembly, unopposed and unanimously, on \textcolor{blue}{\textbf{21 June 2011}}, with effect from 1 January 2012. Mr. Ban describes his priorities as mobilising world leaders to deal with climate change, economic upheaval, pandemics and increasing pressures involving food, energy and water...\\
\hline \textbf{Unfaithful Summary}: The United Nations Secretary-General Ban Ki-moon was elected for a second term in \textcolor{red}{\textbf{2007}}. \\
\hline \textbf{Our Summary}: The United Nations Secretary-General Ban Ki-moon was elected for a second term in \textcolor{blue}{\textbf{21 June 2011}}.\\
\hline
\end{tabular}
\caption{An example \emph{unfaithful} summary.
It suffers from \emph{extrinsic} hallucination, where information not present in the source document was generated.
Our method attempts to correct the unfaithful summary by replacing "2007" with an entity from the source with compatible semantic type (i.e. \textsc{Date}).}
\label{tab:summary_example}
\end{table}
Table \ref{tab:summary_example} shows an example of such summary, generated by BART \cite{lewis2019bart}, an auto-regressive, transformer-based sequence-to-sequence model. 
The article describes an event where the former UN-Secretary-General Ban Ki-Moon was re-elected for a second term. The model hallucinates "2007", which never appears in the source document, leading to inconsistency with the correct date of the event presented. 

In this work, we focus on the problem of \emph{correcting} such hallucinations as a post processing step\footnote{Our code and data is available at \url{http://cogcomp.org/page/publication_view/938}}. 
A post processing correction step allows us to rely on the fluency of the text generated by SOTA systems, that gain from huge pretrained models and large fine-tuning datasets, and correct it using small amounts of automatically generated training data.

Under the setting where a large fraction of ground truth summarization data is hallucinated, 
as we show in Table~\ref{tab:human_error_analysis},
we study the method of \emph{contrast candidate generation} and \emph{selection}. In the generation step, we replace named entities in a potentially hallucinated summary with 
ones with compatible semantic types that are present 
in the source, and create 
variants of candidate summaries. In the selection step, we rank the generated candidates with a discriminative model trained to distinguish between faithful summaries and synthetic negative candidates generated given the source. We experiment 
on a range of RNN- and transformer-based abstractive summarization models. Our preliminary results on the XSum corpus \cite{narayan2018don}, which contains substantial presence of hallucinated ground truth examples, show the effectiveness of our method in correcting unfaithful summaries with \emph{extrinsic hallucinations}. 

Our main contributions are as follows. First, our work is the first to study the effectiveness of \emph{contrast candidate generation} and \emph{selection} as a model-agnostic method for correcting hallucinations, under the setting where a large fraction of ground truth summarization data suffers from hallucinations.
Second, we validate our method on various neural summarization systems trained on XSum, and provide detailed analysis on the typical types of hallucinations from each system.

\section{Contrast Candidate Generation \& Selection}
Our proposed method is built on the observation that a large fraction of extrinsic hallucinations happen on named entities and quantities. Table \ref{tab:human_error_analysis} shows the human analysis by \citet{Maynez2020OnFA} on the hallucinations of $500$ randomly sampled gold summaries from the XSum corpus 
. We break down each category and annotate the proportion of hallucinations that happen on entity and number/quantity spans.

\begin{table}[h]
\centering
\begin{tabular}{lc|cc}
\hline Type & \% & Ent. \% & Num. \%\\
\hline 
Faithful & $23.1$ & - & - \\
Ex. Hallucination & 73.1 & 35.9 & 18.2 \\
In. Hallucination & 7.4 & 1.9 & 0.5\\
\hline
\end{tabular}
\caption{Frequency of extrinsic and intrinsic hallucinations in 500 ground truth summary of the XSum corpus. The ``\%'' column shows the $\%$ of intrinsic and extrinsic hallucinations annotated by \citet{Maynez2020OnFA}. We analyzed the $\%$ of hallucinations on entities and numbers/quantities, and show the $\%$ out of all 500 summaries in the right two columns.}
\label{tab:human_error_analysis}
\end{table}

As \citet{Maynez2020OnFA} further show that the hallucinations in training data translate to similar issues for the generated outputs across different summarization models, we want to study a model-agnostic, post-processing method that can correct such entity and quantity hallucinations. We frame the problem as a correction task and make it conceptually a less complex problem than summarization. Modeling correction as a standalone task would require less training data, which becomes crucial when a large proportion of ground truth summarization data suffer from hallucinations, and inherit the fluency of data intensive SOTA models.


\subsection{Contrast Candidate Generation}
From a model-generated summary, we first identify any potentially hallucinated entities or quantities by checking whether entities with similar surface forms have appeared in the source document.  
We use a neural Named Entity Recognition (NER) system from the Stanza NLP toolkit \cite{qi2020stanza} trained on the OntoNotes corpus \cite{weischedel2013ontonotes} to extract named entities of different semantic types from the source document and summary.  
Each 
named entity present in the summary is replaced
with a different entity present in the document with the same NER label. This gives us different variants of the original summary with the same level of fluency , but not necessarily faithful. 

\subsection{Contrast Candidate Selection}

For the candidate selection step, we want to identify the best candidate among the variants generated in the previous step as the final output summary. As the contrast candidates vary in no more than a few tokens from the original summary, it requires a model with more delicate local decision boundaries  \cite{gardner-etal-2020-evaluating} to select the correct candidate. For example, we observe that MNLI models \cite{williams2017broad} fail to produce satisfactory results.

To create training data for that 
purpose, we sample examples from the XSum training set where all entities in the ground truth summary 
appear in the source document. We then  follow the same procedure in the 
generation step, and produce \emph{unfaithful} variants from the ground truth summary by replacing entities with others that have the same semantic type but
different surface form in the source text. With the ground truth and synthetic negative summaries, we train a text classifier with a discriminative objective to score and rank the variants of the summaries. 

We use BART \cite{lewis2019bart} plus a linear layer as our classification model. We adopt a similar learning objective to contrastive learning \cite{khosla2020supervised}. For each pair of positive and negative summary candidate, we use cross entropy loss $\mathcal{L}_{\textsc{XE}}$ to handle the correctness of the label predictions. We add a margin ranking loss term $\mathcal{L}_{\textsc{Rank}}$ to 
encourage the model to assign higher probability to the positive than the negative candidate. The margin $\gamma$ is a tunable hyperparameter in training.
\begin{align*}
    \mathcal{L} = \mathcal{L}_{\textsc{XE}}(\hat{y_+}, 1) &+ \mathcal{L}_{\textsc{XE}}(\hat{y_-}, 0) + \mathcal{L}_{\textsc{Rank}}(\hat{y_+}, \hat{y_-}) \\
    \mathcal{L}_{\textsc{Rank}} &= max(0, \hat{y_-} - \hat{y_+} + \gamma)
\end{align*}
During test time, we use the trained model to score the generated contrast candidate summaries, as well as the original version generated by the summarization model. We take the candidate with the highest score as the final summary.  
\section{Experiments}

\begin{table}[h]
\centering
\begin{tabular}{l|cc|c}
\hline
\multicolumn{4}{c}{\textit{Full XSum Test Set}} \\
\hline Method & \textsc{Rouge$_L$} & \textsc{ Bert   } & \textsc{Feqa (\%)} \\
\hline 
BART$_{large}$ & \textbf{36.95} & \textbf{91.57} & - \\
+ correct  & 36.70 & 91.50 & - \\
\hline
\multicolumn{4}{c}{\textit{Changed Summary Only (13.3\%)}} \\
\hline
BART$_{large}$ & \textbf{38.63} & \textbf{91.61} & 22.50 \\
+ correct  & 36.62 & 91.10 & \textbf{25.62} \\
\hline
\end{tabular}
\caption{Evaluation with automatic metrics on the summaries generated by the baseline BART$_{large}$ model, plus our post-processing correction method. We report $F_{\beta=1}$ scores with \textsc{Rouge} and \textsc{BertScore}, plus the macro-averaged percentage of questions answered correctly for each summary with \textsc{Feqa}, a QA-based metric for summary faithfulness proposed by \citet{durmus-etal-2020-feqa}.  }
\label{tab:rouge-results}
\end{table}

Our experiments focus on the aforementioned XSum corpus, where the target summary is highly \emph{abstractive} and likely \emph{hallucinated}. We first consider the summaries generated by a BART model trained on the XSum corpus. 
By applying our method, we are able to change $13.3\%$ of all model generated summaries. For $38.4\%$ of all summaries, the original summary does not have a hallucinated entity, or there is no entity with compatible type in the source text. 
Our model decides to keep the original summary in the rest $48.3\%$.  

\subsection{\textsc{Rouge} and \textsc{BertScore} Evaluation}
We first verify that our method does not hurt the fluency and salience of the generated summaries, for which we assume \textsc{Rouge} \cite{lin-2004-rouge} and \textsc{BertScore} \cite{bert-score} are suitable metrics. 
We report the results in Table~\ref{tab:rouge-results}. 
We observe though both the baseline and our method do well in both \textsc{Rouge} and \textsc{BertScore}, our method trails behind in both metrics slightly. This is due to the existence of extrinsic hallucinations in the ground truth summary, and the model manages to generate a part of the hallucinations, and gets incorrectly rewarded by such.

\subsection{Faithfulness Evaluation}

\begin{table}[t]
\centering
\begin{tabular}{lccc}
\hline Method & \textit{Faith. \%} & \textit{Ex. \%} & \textit{In. \%}\\
\hline 
\small{BART} & 23.8$\pm{9.6}$ & 71.7$\pm{11.2}$ & \textbf{1.7}$\pm{3.5}$   \\
\small{+ correct} & \textbf{59.5}$\pm{12.4}$ & \textbf{9.2}$\pm{7.3}$  & 29.1$\pm{11.6}$ \\
\hline
\end{tabular}
\caption{Percentage of examples human annotator judged as ``faithful'' (\textit{Faith.}), ``extrinsically hallucinated'' (\textit{Ex.}), and ``intrinsically hallucinated'' (\textit{In.}) among the 95 randomly sampled  
corrected summaries. The $95\%$ confidence intervals are estimated with bootstrap resampling (Appendix. \ref{sec:human_eval_appendix}).}
\label{tab:bart_human_eval}
\end{table}

To test whether our correction method can improves the faithfulness of the summaries, we evaluate the summaries with \textsc{Feqa} \cite{durmus-etal-2020-feqa}, a QA-based metric for summary faithfulness. Given a summary, \textsc{Feqa} automatically generates questions on noun phrase and named entity spans in the summary, and uses a pretrained QA model to verify if the answer derived from the source document exact-matches the span in the summary. 

We run \textsc{Feqa} and compute the macro-averaged percentage of questions answered correctly for each of the $1510$ summaries that our system made corrections to, and report the results in Table~\ref{tab:rouge-results}. The results suggest that the corrected summaries present statistically significant improvements over the original ones ($p<0.001$, with a two-tailed, paired t-test).
\begin{table*}[t]
\small
\centering
\begin{tabular}{p{0.18\linewidth}|p{0.1\linewidth}|p{0.65\linewidth}}
\hline
\multicolumn{3}{c}{\textit{Good Corrections}} \\
\hline
Type & System & Original Summary and Our Change \\
\hline 
{\small Correcting NE Hallucination} & \textsc{BertS2S} & {\small Tranmere Rovers have signed midfielder \textcolor{red}{[Alfreton]$_{\textsc{PER}}$} $\rightarrow$ \textcolor{blue}{[Mooney]$_{\textsc{PER}}$} on loan until the end of the season.}   \\
{\small Correcting Number Hallucination} & \textsc{Bart} & {\small A judge has ruled that the \textcolor{red}{[\$9.6bn (£5.03bn)]$_{\textsc{Money}}$} $\rightarrow$ \textcolor{blue}{[\$7.8bn (£5.03bn)]$_{\textsc{Money}}$} oil spill compensation fund is not fraudulent.} \\
\hline
\multicolumn{3}{c}{\textit{Typical Mistakes}} \\
\hline
{\small No correct replacement exists in source} & \textsc{Bart} & {\small Helmut Kohl, who has died at the age of \textcolor{red}{[87]$_{\textsc{Cardinal}}$} $\rightarrow$ \textcolor{orange}{[39]$_{\textsc{Cardinal}}$}, was one of the driving forces behind Germany's reunification in 1990.}\\
{\small Wrong type of NE in summary} & \textsc{TranS2S} & {\small \textcolor{red}{[Andrew Marr]$_{\textsc{PER}}$} $\rightarrow$ \textcolor{orange}{[Venter]$_{\textsc{PER}}$} is one of the most important scientific discoveries in human life.} \\
{\small Not explicit in source, but can be inferred} & \textsc{BertS2S} & {\small \textcolor{blue}{[Three]$_{\textsc{Cardinal}}$} $\rightarrow$ \textcolor{orange}{[Two]$_{\textsc{Cardinal}}$} fugitives have been arrested and charged with attempting to smuggle drugs into the country.} \\
\hline
\end{tabular}
\caption{Examples of corrections and typical mistakes made by our proposed method on generated summaries by different summarization models. The original and replaced entities in each summary are highlighted, and are colored by their faithfulness categories (\textcolor{red}{Red}: \textit{Extrinsic Hallucation}; \textcolor{orange}{Orange}: \textit{Intrinsic Hallucation}; \textcolor{blue}{Blue}: \textit{Faithful}) }
\label{tab:examples}
\end{table*}

\begin{table}[h]
\centering
\begin{tabular}{l|ccc|c}
\hline System & \textsc{$P$} & \textsc{$R$} & \textsc{$F_1$} & \textsc{Ent. \%} \\
\hline 
\textsc{PtGen}  & 79.86 & 58.38 & 67.45 & 65.48  \\
\textsc{TConvS2S}  & 87.76 & 61.87 & 72.57 & 64.27 \\
\textsc{TranS2S}  & 81.81 & 57.35 & 67.44 & 80.15 \\
\textsc{BertS2S}  & 80.54 & 37.82 & 51.47 & 56.85 \\
\hline
\end{tabular}
\caption{The selection model's precision, recall and $F_{1}$ on identifying hallucinated output from four different summarization systems. 
The \textsc{Ent. \%} column shows the $\%$ of hallucinations on entities and quantities among all hallucinated summaries by each system. }
\label{tab:cross-system}
\end{table}

Table~\ref{tab:bart_human_eval} shows the human evaluation results on the 95 randomly sampled subset of changed summaries. Two expert annotators assign each summary into three faithfulness categories and adjudicate the decisions. Additional annotations from a third expert is then used to calculate the inter-annotator agreement. 
As the results show, our model is able to improve the faithfulness of the summaries, but at the cost of incurring intrinsic hallucinations on mistakes, which we will discuss more in detail in section \ref{ssec:trade_off}.

\section{Analysis and Discussion}

\subsection{Identifying Hallucination Across Systems}
Table~\ref{tab:cross-system} shows our selection model's performance when measuring P, R, F$_1$ w.r.t all the hallucinated instances. 
We use the test set from \citet{Maynez2020OnFA}, who have annotated hallucination categories of generated summaries from four neural summariazaiton models: \textsc{PtGen} \cite{see2017get} \textsc{TConvS2S} \cite{narayan2018don}, \textsc{BertS2S} and \textsc{TranS2S} \cite{rothe2020leveraging}. 
Our system achieves consistently high level of precision across models. The system achieves high relative recall with respect to the $\%$ of entity and quantity hallucinations among all hallucinations. As our method only targets entities and quantities, the overall recall varies by the typical type of hallucinations each summarization system makes.
We also observe while our method achieves high recall on models with lower \textsc{Rouge} and \textsc{BertScore}, the recall drops on pretrained models such as \textsc{BertS2S}. This is potentially due to the decreased percentage of entity/quantity hallucinations exist in generated summaries from the models with pretraining. 



\subsection{Intrinsic vs. Extrinsic Hallucinations Trade-off} \label{ssec:trade_off}
As our method detects and corrects extrinsic-hallucinated entities, naturally any entities replaced wrong would introduce intrinsic hallucinations in the changed summary, as indicated by the results in Table~\ref{tab:bart_human_eval}. To speculate why the mistakes happen, we analyzed the typical mistakes by the model, and listed a few representative examples in Table~\ref{tab:examples}. 
For example, our method could not find the correct replacement for a hallucinated entity when no such one exists in the source text. We observe that the models with pretraining, such as \textsc{BertS2S}, \cite{rothe2020leveraging} and \textsc{Bart}, suffer from the issue by most, as they tend to be affected by artifacts/priors from the pretraining process. 
\subsection{Entity Faithfulness $\subsetneqq $ Summary Faithfulness}
From the observation that models often hallucinate entities with no correct replacement in the source, we suspect that solving entity faithfulness alone does not guarantee the faithfulness of the summary. In the last example from Table~\ref{tab:examples}, the \textsc{BertS2S} system correctly identifies that three fugitives are involved in the event described by the source text, even though the number "three" has never been explicitly mentioned in the source context in any surface forms. Furthermore, statistics provided by \citet{Maynez2020OnFA} show that abstractive summarization models often produces \emph{factual} statements, i.e. verifiable in the real world independent of the source text. Such findings imply that identifying hallucinations often requires more complex objectives such as commonsense reasoning and knowledge retrieval. The solution we propose here that focuses only on entites and quantities would likely be insufficient to solve the entire problem.

\section{Related Work}
There have been growing interests in quantitatively measuring the faithfulness of text generation models. Most widely-adopted evaluation metrics for text generation, such as ROUGE \cite{lin-2004-rouge} and BERTScore \cite{bert-score}, correlate poorly with the human perceived faithfulness of the generated text \cite{kryscinski2019neural, durmus-etal-2020-feqa}. 
Recent studies explore categorical, content-based analysis for measuring the faithfulness of summaries \cite{goyal-durrett-2020-evaluating, deutsch2020understanding}. \citet{narayan-etal-2018-ranking, deutsch2020towards, durmus-etal-2020-feqa} propose to use question answering to test the consistency of summary content to the information presented in the source text.  

 
There have been efforts to study pre- or post- processing methods to improving faithfulness of generated summaries. \citet{falke-etal-2019-ranking} attempt to use textual entailment models to re-rank the summary candidates generated from beam search or different neural systems. As \citet{Maynez2020OnFA} highlight the existence of hallucinations in training data, truncating potentially unfaithful gold summaries during training is an effective strategy \cite{Kang2020ImprovedNL, filippova-2020-controlled}. \citet{kryscinski2020evaluating} take similar apporach as in this work to identify the hallucinations in summary. 
A concurrent study to this work \cite{cao2020factual} uses similar strategies as in this paper on a dataset with a very small fraction of hallucinations present. Our study instead focuses on the more challenging setting \cite{goyal2020annotating} where a large part of training data suffers from extrinsic and intrinsic hallucinations, and provides cross-system analysis on the both hallucinations categories.

\section{Conclusion}
We study contrast candidate generation and selection as a method to apply post-hoc fixes to extrinsically hallucinated summary on entities and quantities, under the setting where the summarization dataset suffers from intrinsic and extrinsic hallucinations. We conduct our experiments on the XSum dataset, and show that our method is able to correct extrinsic hullucinations, but incurs a small fraction of intrinsic hallucinations on mistakes. We also provide detailed analysis and discussions on the capabilities and limitations of our method. We hope our findings in the paper will provide insights to future work in this direction.

\section*{Acknowledgments}
We thank Sunita Verma and Sugato Basu for valuable input and feedback on drafts of the paper. 
This work was supported in part by a Focused Award from Google, a gift from Tencent, and by Contract FA8750-19-2-1004 with the US Defense Advanced Research Projects Agency (DARPA). The views expressed are those of the authors and do not reflect the official policy or position of the Department of Defense or the U.S. Government.
\bibliographystyle{acl_natbib}
\bibliography{ccg, cited_00s, cited_10s, cited_90s, new}

\begin{thebibliography}{29}
\expandafter\ifx\csname natexlab\endcsname\relax\def\natexlab#1{#1}\fi

\bibitem[{Cao et~al.(2020)Cao, Dong, Wu, and Cheung}]{cao2020factual}
Meng Cao, Yue Dong, Jiapeng Wu, and Jackie Chi~Kit Cheung. 2020.
\newblock \href {https://doi.org/10.18653/v1/2020.emnlp-main.506} {Factual
  error correction for abstractive summarization models}.
\newblock In \emph{Proceedings of the 2020 Conference on Empirical Methods in
  Natural Language Processing (EMNLP)}, pages 6251--6258, Online. Association
  for Computational Linguistics.

\bibitem[{Deutsch et~al.(2020)Deutsch, Bedrax-Weiss, and
  Roth}]{deutsch2020towards}
Daniel Deutsch, Tania Bedrax-Weiss, and Dan Roth. 2020.
\newblock Towards question-answering as an automatic metric for evaluating the
  content quality of a summary.
\newblock \emph{arXiv preprint arXiv:2010.00490}.

\bibitem[{Deutsch and Roth(2020)}]{deutsch2020understanding}
Daniel Deutsch and Dan Roth. 2020.
\newblock Understanding the extent to which summarization evaluation metrics
  measure the information quality of summaries.
\newblock \emph{arXiv preprint arXiv:2010.12495}.

\bibitem[{Devlin et~al.(2019)Devlin, Chang, Lee, and
  Toutanova}]{devlin2019bert}
Jacob Devlin, Ming-Wei Chang, Kenton Lee, and Kristina Toutanova. 2019.
\newblock \href {https://doi.org/10.18653/v1/N19-1423} {{BERT}: Pre-training of
  deep bidirectional transformers for language understanding}.
\newblock In \emph{Proceedings of the 2019 Conference of the North {A}merican
  Chapter of the Association for Computational Linguistics: Human Language
  Technologies, Volume 1 (Long and Short Papers)}, pages 4171--4186,
  Minneapolis, Minnesota. Association for Computational Linguistics.

\bibitem[{Durmus et~al.(2020)Durmus, He, and Diab}]{durmus-etal-2020-feqa}
Esin Durmus, He~He, and Mona Diab. 2020.
\newblock \href {https://doi.org/10.18653/v1/2020.acl-main.454} {{FEQA}: A
  question answering evaluation framework for faithfulness assessment in
  abstractive summarization}.
\newblock In \emph{Proceedings of the 58th Annual Meeting of the Association
  for Computational Linguistics}, pages 5055--5070, Online. Association for
  Computational Linguistics.

\bibitem[{Falke et~al.(2019)Falke, Ribeiro, Utama, Dagan, and
  Gurevych}]{falke-etal-2019-ranking}
Tobias Falke, Leonardo F.~R. Ribeiro, Prasetya~Ajie Utama, Ido Dagan, and Iryna
  Gurevych. 2019.
\newblock \href {https://doi.org/10.18653/v1/P19-1213} {Ranking generated
  summaries by correctness: An interesting but challenging application for
  natural language inference}.
\newblock In \emph{Proceedings of the 57th Annual Meeting of the Association
  for Computational Linguistics}, pages 2214--2220, Florence, Italy.
  Association for Computational Linguistics.

\bibitem[{Filippova(2020)}]{filippova-2020-controlled}
Katja Filippova. 2020.
\newblock \href {https://www.aclweb.org/anthology/2020.findings-emnlp.76}
  {Controlled hallucinations: Learning to generate faithfully from noisy data}.
\newblock In \emph{Findings of the Association for Computational Linguistics:
  EMNLP 2020}, pages 864--870, Online. Association for Computational
  Linguistics.

\bibitem[{Gardner et~al.(2020)Gardner, Artzi, Basmov, Berant, Bogin, Chen,
  Dasigi, Dua, Elazar, Gottumukkala, Gupta, Hajishirzi, Ilharco, Khashabi, Lin,
  Liu, Liu, Mulcaire, Ning, Singh, Smith, Subramanian, Tsarfaty, Wallace,
  Zhang, and Zhou}]{gardner-etal-2020-evaluating}
Matt Gardner, Yoav Artzi, Victoria Basmov, Jonathan Berant, Ben Bogin, Sihao
  Chen, Pradeep Dasigi, Dheeru Dua, Yanai Elazar, Ananth Gottumukkala, Nitish
  Gupta, Hannaneh Hajishirzi, Gabriel Ilharco, Daniel Khashabi, Kevin Lin,
  Jiangming Liu, Nelson~F. Liu, Phoebe Mulcaire, Qiang Ning, Sameer Singh,
  Noah~A. Smith, Sanjay Subramanian, Reut Tsarfaty, Eric Wallace, Ally Zhang,
  and Ben Zhou. 2020.
\newblock \href {https://www.aclweb.org/anthology/2020.findings-emnlp.117}
  {Evaluating models{'} local decision boundaries via contrast sets}.
\newblock In \emph{Findings of the Association for Computational Linguistics:
  EMNLP 2020}, pages 1307--1323, Online. Association for Computational
  Linguistics.

\bibitem[{Goyal and Durrett(2020)}]{goyal-durrett-2020-evaluating}
Tanya Goyal and Greg Durrett. 2020.
\newblock \href {https://www.aclweb.org/anthology/2020.findings-emnlp.322}
  {Evaluating factuality in generation with dependency-level entailment}.
\newblock In \emph{Findings of the Association for Computational Linguistics:
  EMNLP 2020}, pages 3592--3603, Online. Association for Computational
  Linguistics.

\bibitem[{Goyal and Durrett(2021)}]{goyal2020annotating}
Tanya Goyal and Greg Durrett. 2021.
\newblock Annotating and modeling fine-grained factuality in summarization.
\newblock In \emph{Proceedings of the 2021 Proceedings Conference of the North
  {A}merican Chapter of the Association for Computational Linguistics: Human
  Language Technologies.}

\bibitem[{Kang and Hashimoto(2020)}]{Kang2020ImprovedNL}
Daniel Kang and Tatsunori Hashimoto. 2020.
\newblock \href {https://doi.org/10.18653/v1/2020.acl-main.66} {Improved
  natural language generation via loss truncation}.
\newblock In \emph{Proceedings of the 58th Annual Meeting of the Association
  for Computational Linguistics}, pages 718--731, Online. Association for
  Computational Linguistics.

\bibitem[{Khosla et~al.(2020)Khosla, Teterwak, Wang, Sarna, Tian, Isola,
  Maschinot, Liu, and Krishnan}]{khosla2020supervised}
Prannay Khosla, Piotr Teterwak, Chen Wang, Aaron Sarna, Yonglong Tian, Phillip
  Isola, Aaron Maschinot, Ce~Liu, and Dilip Krishnan. 2020.
\newblock Supervised contrastive learning.
\newblock \emph{{Proceedings of the 34th Conference on Neural Information
  Processing Systems}}.

\bibitem[{Kryscinski et~al.(2019)Kryscinski, Keskar, McCann, Xiong, and
  Socher}]{kryscinski2019neural}
Wojciech Kryscinski, Nitish~Shirish Keskar, Bryan McCann, Caiming Xiong, and
  Richard Socher. 2019.
\newblock \href {https://doi.org/10.18653/v1/D19-1051} {Neural text
  summarization: A critical evaluation}.
\newblock In \emph{Proceedings of the 2019 Conference on Empirical Methods in
  Natural Language Processing and the 9th International Joint Conference on
  Natural Language Processing (EMNLP-IJCNLP)}, pages 540--551, Hong Kong,
  China. Association for Computational Linguistics.

\bibitem[{Kryscinski et~al.(2020)Kryscinski, McCann, Xiong, and
  Socher}]{kryscinski2020evaluating}
Wojciech Kryscinski, Bryan McCann, Caiming Xiong, and Richard Socher. 2020.
\newblock Evaluating the factual consistency of abstractive text summarization.
\newblock In \emph{Proceedings of the 2020 Conference on Empirical Methods in
  Natural Language Processing (EMNLP)}, pages 9332--9346.

\bibitem[{Lewis et~al.(2020)Lewis, Liu, Goyal, Ghazvininejad, Mohamed, Levy,
  Stoyanov, and Zettlemoyer}]{lewis2019bart}
Mike Lewis, Yinhan Liu, Naman Goyal, Marjan Ghazvininejad, Abdelrahman Mohamed,
  Omer Levy, Veselin Stoyanov, and Luke Zettlemoyer. 2020.
\newblock \href {https://doi.org/10.18653/v1/2020.acl-main.703} {{BART}:
  Denoising sequence-to-sequence pre-training for natural language generation,
  translation, and comprehension}.
\newblock In \emph{Proceedings of the 58th Annual Meeting of the Association
  for Computational Linguistics}, pages 7871--7880, Online. Association for
  Computational Linguistics.

\bibitem[{Lin(2004)}]{lin-2004-rouge}
Chin-Yew Lin. 2004.
\newblock \href {https://www.aclweb.org/anthology/W04-1013} {{ROUGE}: A package
  for automatic evaluation of summaries}.
\newblock In \emph{Text Summarization Branches Out}, pages 74--81, Barcelona,
  Spain. Association for Computational Linguistics.

\bibitem[{Liu and Lapata(2019)}]{liu2019text}
Yang Liu and Mirella Lapata. 2019.
\newblock Text summarization with pretrained encoders.
\newblock In \emph{Proceedings of the 2019 Conference on Empirical Methods in
  Natural Language Processing and the 9th International Joint Conference on
  Natural Language Processing (EMNLP-IJCNLP)}, pages 3721--3731.

\bibitem[{Maynez et~al.(2020)Maynez, Narayan, Bohnet, and
  McDonald}]{Maynez2020OnFA}
Joshua Maynez, Shashi Narayan, Bernd Bohnet, and Ryan McDonald. 2020.
\newblock \href {https://doi.org/10.18653/v1/2020.acl-main.173} {On
  faithfulness and factuality in abstractive summarization}.
\newblock In \emph{Proceedings of the 58th Annual Meeting of the Association
  for Computational Linguistics}, pages 1906--1919, Online. Association for
  Computational Linguistics.

\bibitem[{Nallapati et~al.(2016)Nallapati, Zhou, dos Santos,
  GuÌ‡l{\c{c}}ehre, and Xiang}]{nallapati2016abstractive}
Ramesh Nallapati, Bowen Zhou, Cicero dos Santos, {\c{C}}a{\u{g}}lar
  GuÌ‡l{\c{c}}ehre, and Bing Xiang. 2016.
\newblock \href {https://doi.org/10.18653/v1/K16-1028} {Abstractive text
  summarization using sequence-to-sequence {RNN}s and beyond}.
\newblock In \emph{Proceedings of The 20th {SIGNLL} Conference on Computational
  Natural Language Learning}, pages 280--290, Berlin, Germany. Association for
  Computational Linguistics.

\bibitem[{Narayan et~al.(2018{\natexlab{a}})Narayan, Cohen, and
  Lapata}]{narayan2018don}
Shashi Narayan, Shay~B Cohen, and Mirella Lapata. 2018{\natexlab{a}}.
\newblock Don’t give me the details, just the summary! topic-aware
  convolutional neural networks for extreme summarization.
\newblock In \emph{Proceedings of the 2018 Conference on Empirical Methods in
  Natural Language Processing}, pages 1797--1807.

\bibitem[{Narayan et~al.(2018{\natexlab{b}})Narayan, Cohen, and
  Lapata}]{narayan-etal-2018-ranking}
Shashi Narayan, Shay~B. Cohen, and Mirella Lapata. 2018{\natexlab{b}}.
\newblock \href {https://doi.org/10.18653/v1/N18-1158} {Ranking sentences for
  extractive summarization with reinforcement learning}.
\newblock In \emph{Proceedings of the 2018 Conference of the North {A}merican
  Chapter of the Association for Computational Linguistics: Human Language
  Technologies, Volume 1 (Long Papers)}, pages 1747--1759, New Orleans,
  Louisiana. Association for Computational Linguistics.

\bibitem[{Qi et~al.(2020)Qi, Zhang, Zhang, Bolton, and Manning}]{qi2020stanza}
Peng Qi, Yuhao Zhang, Yuhui Zhang, Jason Bolton, and Christopher~D. Manning.
  2020.
\newblock \href {https://nlp.stanford.edu/pubs/qi2020stanza.pdf} {Stanza: A
  {Python} natural language processing toolkit for many human languages}.
\newblock In \emph{Proceedings of the 58th Annual Meeting of the Association
  for Computational Linguistics: System Demonstrations}.

\bibitem[{Rothe et~al.(2020)Rothe, Narayan, and Severyn}]{rothe2020leveraging}
Sascha Rothe, Shashi Narayan, and Aliaksei Severyn. 2020.
\newblock Leveraging pre-trained checkpoints for sequence generation tasks.
\newblock \emph{Transactions of the Association for Computational Linguistics},
  8:264--280.

\bibitem[{Rush et~al.(2015)Rush, Chopra, and Weston}]{rush2015neural}
Alexander~M Rush, Sumit Chopra, and Jason Weston. 2015.
\newblock A neural attention model for abstractive sentence summarization.
\newblock In \emph{Proceedings of the 2015 Conference on Empirical Methods in
  Natural Language Processing}, pages 379--389.

\bibitem[{See et~al.(2017)See, Liu, and Manning}]{see2017get}
Abigail See, Peter~J Liu, and Christopher~D Manning. 2017.
\newblock Get to the point: Summarization with pointer-generator networks.
\newblock In \emph{Proceedings of the 55th Annual Meeting of the Association
  for Computational Linguistics (Volume 1: Long Papers)}, pages 1073--1083.

\bibitem[{Vaswani et~al.(2017)Vaswani, Shazeer, Parmar, Uszkoreit, Jones,
  Gomez, Kaiser, and Polosukhin}]{vaswani2017attention}
Ashish Vaswani, Noam Shazeer, Niki Parmar, Jakob Uszkoreit, Llion Jones,
  Aidan~N Gomez, {\L}ukasz Kaiser, and Illia Polosukhin. 2017.
\newblock Attention is all you need.
\newblock In \emph{Advances in Neural Information Processing Systems}, pages
  5998--6008.

\bibitem[{Weischedel et~al.(2013)Weischedel, Palmer, Marcus, Hovy, Pradhan,
  Ramshaw, Xue, Taylor, Kaufman, Franchini et~al.}]{weischedel2013ontonotes}
Ralph Weischedel, Martha Palmer, Mitchell Marcus, Eduard Hovy, Sameer Pradhan,
  Lance Ramshaw, Nianwen Xue, Ann Taylor, Jeff Kaufman, Michelle Franchini,
  et~al. 2013.
\newblock Ontonotes release 5.0.
\newblock \emph{Linguistic Data Consortium, Philadelphia, PA}, 23.

\bibitem[{Williams et~al.(2018)Williams, Nangia, and
  Bowman}]{williams2017broad}
Adina Williams, Nikita Nangia, and Samuel Bowman. 2018.
\newblock \href {https://doi.org/10.18653/v1/N18-1101} {{A Broad-Coverage
  Challenge Corpus for Sentence Understanding through Inference}}.
\newblock In \emph{Proceedings of the 2018 Conference of the North {A}merican
  Chapter of the Association for Computational Linguistics: Human Language
  Technologies, Volume 1 (Long Papers)}, pages 1112--1122, New Orleans,
  Louisiana. Association for Computational Linguistics.

\bibitem[{Zhang et~al.(2020)Zhang, Kishore, Wu, Weinberger, and
  Artzi}]{bert-score}
Tianyi Zhang, Varsha Kishore, Felix Wu, Kilian~Q. Weinberger, and Yoav Artzi.
  2020.
\newblock \href {https://openreview.net/forum?id=SkeHuCVFDr} {{BERTScore:
  Evaluating Text Generation with BERT}}.
\newblock In \emph{International Conference on Learning Representations}.

\end{thebibliography}

\newpage
\appendix
\section{Candidate Selection Model}
For our contrast candidate selection model, we use a pretrained BART base model. We add a linear layer over the max pooled embedding, and the classification model is expected to output a label between ["\textsc{Faithful}", "\textsc{Hallucinated}"]. \\
For all our experiments, we use the following set of hyper-parameters: $r=1e-5$, margin $\gamma = 0$, number of training epoch$=3$.
\section{Complete \textsc{Rouge} and \textsc{BertScore} Results}
\label{sec:rouge_eval}

\begin{table}[h]
\centering
\begin{tabular}{lcccc}
\hline
\multicolumn{5}{c}{\textit{Full XSum Test Set}} \\
\hline Method & \textsc{$R_1$} & \textsc{$R_2$} & \textsc{$R_L$} & \textsc{Bert} \\
\hline 
BART$_{large}$ & 45.10 & 21.86 & 36.95 & 91.57  \\
+ correct  & 44.82 & 21.49 & 36.70 & 91.50 \\
\hline
\multicolumn{5}{c}{\textit{Changed Summary Only (13.3\%)}} \\
\hline
BART & 46.73 & 23.51 & 38.63 & 91.61  \\
+ correct  & 44.35 & 20.70 & 36.62 & 91.10 \\
\hline
\end{tabular}
\caption{\textsc{Rouge}$_{\{1,2,L\}}$ and \textsc{BertScore} evaluation results (in $F_1$) of summaries generated by the baseline BART$_{large}$ model, plus the corrected summaries with our post-processing method, on the test set of the XSum corpus. }
\label{tab:complete-rouge-results}
\end{table}

\section{Estimating Confidence Interval for Human Evaluation}
\label{sec:human_eval_appendix}
We use bootstrapping to estimate the confidence interval for the expert annotation presented in Table~\ref{tab:bart_human_eval}. For each faithfulness category on the two systems, we regard the adjudicated annotation as ground truth, and label the individual instance as the true positive (\textsc{TP}), false negative (\textsc{FN}), true negative (\textsc{TN}) and false positive (\textsc{FP}) respectively according the annotations from the third expert. We re-sample the 95 instances with replacement for 1,000 times. We estimate the adjusted mean and $95\%$ confidence interval from the mean and standard deviation of the sampled distribution of (\textsc{TP} + \textsc{FN}).

\end{document}